# FREGAN: Frame Rate Enhancement in Videos using Generative Adversarial Networks

Rishik Mishra

Department of Computer Engineering and Applications,
GLA University, Mathura, Uttar Pradesh, India
E-mail: rishikmishra1999@gmail.com

Neeraj Gupta

Department of Computer Engineering and Applications,
GLA University, Mathura, Uttar Pradesh, India
E-mail: neeraj.gupta@gla.ac.in

Nitya Shukla

Department of Computer Engineering and Applications,
GLA University, Mathura, Uttar Pradesh, India
E-mail: nityashukla975@gmail.com

**Abstract**: A digital video is a collection of individual frames, while streaming the video the scene utilized the time slice for each frame. High refresh rate and high frame rate is the demand of all high technology applications. The action tracking in videos becomes easier and motion becomes smoother in gaming applications due to the high refresh rate. It provides a faster response because of less time in between each frame that is displayed on the screen. FREGAN (Frame Rate Enhancement Generative Adversarial Network) model has been proposed, which predicts future frames of a video sequence based on a sequence of past frames. In this paper, we investigated the GAN model and proposed FREGAN for the enhancement of frame rate in videos. We have utilized Huber loss as a loss function in the proposed FREGAN. It provided excellent results in super-resolution and we have tried to reciprocate that performance in the application of frame rate enhancement. We have validated the effectiveness of the proposed model on the standard datasets (UCF101 and RFree500). The experimental outcomes illustrate that the proposed model has a Peak signal-to-noise ratio (PSNR) of 34.94 and a Structural Similarity Index (SSIM) of 0.95.

# 1 Introduction

The refresh rate is defined as the number of times your system refreshes itself every second and this determines how often that information is cycled through your display. While the frame rate is defined as the number of frames sent to the screen every second. The increase in the high refresh rate monitors has paved a way for smartphone's and other cameras to record videos in a high frame rate and to view it at the same level, this is where the industry still lacks, apart from some high-end devices, many other devices only provide the ability to record videos in 60fps or the best case 120fps, while the monitors now can display at 240hz and even more [reference to report on monitors and cameras]. Another field where a high frame rate is utilized extensively is slow-motion video capture, again due to the limited processing power of mobile devices the slow-motion videos are limited to just a few seconds of footage. Predicting future frames enables a high frame rate to make smart decisions in various tasks. An approach towards frame rate enhancement is Frame interpolation, it is the action of generating a frame for a video, given the immediate frames occurring sequentially before and after. This allows a video to have its frame rate enhanced, which is a process known as upsampling. In general upsampling, we can't assume access to the ground truth for the interpolated frames. But the problem with this method is blurry interpolated images do not fit with the overall style of the videos, and are easily detected by the human eye. Multiple methods have proven that Generative Adversarial Networks(GANs)[2] perform relatively better in the study of frame representation learning.

We study this problem and predict a new frame and overcome the drawbacks, proposing a model which uses Generative Adversarial Networks(GAN) to predict frames between two consecutive frames, and with this, we were able to increase the total frame rate while only using 2 frames instead of a sequence of frames. We use an architecture based on the Pix2Pix GAN [19] to increase the frame rate by predicting an intermediate frame given two adjacent frames. We employ a custom Convolutional Neural Network (CNN) model in the discriminator and a modified version of UNet[18] as the generator in the proposed model. For upscaling we use the transpose convolution layer and LeakyReLU activation. This study also proposes the implementation of the Huber loss function for the GAN to improve the performance of the model and for a better representation of the training performance. Being model-based the proposed method gives a sufficiently generalized model.

The main contribution of this paper is that the frame of the videos has been increased by using the FREGAN model and also we have achieved better results in comparison with other methods. The organization of this paper is as follows. In Section 2, we have described the related work in chronological order of the techniques proposed by the previous paper. In Section 3, we have discussed the FREGAN model in detail and the standard dataset that is UCF101[20] and RFree500 that we have used in the proposed model. Section 4, shows the results and discussion of the experimental results. And at last, we concluded all the things in Section 5.

# 2 Related works

Frame rate enhancement is the need of all high-tech gaming applications and auto driving. A lot of previous work has been done on frame rate enhancement and next frame prediction using interpolation of frames, using deep neural networks, and using generative adversarial networks.

## 2.1 Frame Interpolation

The frame interpolation technique is used to increase the frame rate of the videos. In this technique, intermediate frames are generated between the present frames using interpolation. It helps in making the animations more fluid, compensating the blur within the motion, and the effect of fake slow motion and thus video streaming is benefited greatly using this method. In 1999, David Gibson and Michael Spann worked on the Multiple frame estimation. They stated in their paper[1] that the movement trajectories offer a localized description of movement over a pair of frames. They evolved a method that is used withinside the estimation of a spatially dispersed set of movement trajectories to clarify the movement present inside an image. Motion-compensated prediction is the main objective of their method. In 2017[3] Reza Mahjourian and their team researched how geometry helps in the prediction of frames in monocular videos. They stated that they use the anticipated depth from the video input for generating the subsequent frame prediction using scene geometry. The depth is predicted by training the recurrent convolutional neural network. This approach can predict the rich next frame along with the depth information which is attached to each pixel. In this year, Simon Niklaus[4] with their team also introduced a method of interpolation of frames to increase the frame rate.

## 2.2 Deep Neural Network-based methods

Using the frame interpolation technique we achieve good video streaming and also it gives a high frame rate, but in this approach, the generated images are blurry and look like averages of existing frames due to linear interpolation. But in 2D animated videos, this technique was not well fitted. Because these videos have sharp and defined individual frames and blurry interpolated images do not fit within the videos, and these are easily detected by the human eye. When frame interpolation techniques individually are not able to meet the expectation, deep learning is used to enhance the quality of interpolated frames. And by using this technique the frame rate of 2D animated videos increases by generating high quality of frames which are smoothly fitted to these videos and also not detected by the human eye easily. In the year 2018[7], Boxuan Yue and Jun Liang used a deep learning technique for frame prediction. They proposed a new recurrent encoder and a deconvolutional decoder to predict frames in their paper. They introduced residual learning to solve the gradient issues. This method can predict frames fast and in an efficient manner.

## 2.3 GAN technique

In 2014[2], Generative adversarial Networks was introduced by IanJ. Goodfellow and their teams. And using this technique, In 2019[9] Sandra Aigner and Marco Körner introduced Future GAN to predict the Future Frames of Video Sequences using Spatio-Temporal 3d Convolutions in Progressively Growing GANs. They proposed a new GAN architecture that predicts future frames based on sequential past frames. In this year Wei Xue[8] and their team also worked on the GAN technique and introduced a new kind of GAN named FrameGAN to increase the frame rate of gait videos. They stated that the FrameGAN will reduce the gap between adjacent frames to increase the frame rate. They also proposed an effective loss function named Margin Ratio Loss (MRL) to improve the recognition model. In 2020[10] Wang Shen and their team worked on BlurryVideo Frame Interpolation. They proposed a method to reduce motion blur and upconvert frame rate simultaneously. Their method performs against state-of-the-art methods. In this year many methods[12][14] used deep learning techniques to enhance the frame rate of videos.

## 3 Proposed Model

In this section, we discuss the method that we used to increase the frame rate and the dataset on which we have experimented. We introduce the FREGAN model, which consists of a generator and a discriminator model with their loss function. We have applied the pseudo-Huber loss function for a more robust prediction. We trained this model on 2 datasets that are UCF101[20] and RFree500.

### 3.1 FREGAN Model

The FREGAN model is based on the Pix2Pix model and thus consists of 2 models, a generator and a discriminator with the generator network trained to predict an intermediate frame given a pair of subsequent frames, and the discriminator networks are trained to differentiate between a true frame and a fake/predicted frame as can be seen from algorithm 1.

---

**Algorithm 1** FREGAN, the proposed algorithm, after finding the optimum value of delta parameter to be 0.5, with dataset1, dataset2, dataset3 being the pre-divided dataset with consecutive frames.

---

**Input** : Sequential Frames of a video, arranged as x1,x2,x3, with x1 being the first sequential frame and x3 being the third
**Output** : frame $X_2$' (ie. FREGAN($X_1,X_2$))
1. **While** $\Theta_g$ and $\Theta_d$ does not converge **do**
2.   **For** step in {0,...,steps} **do**
3.     real batch[x_real1, x_real2, x_real3]=dataset1[0:size],dataset2[0:size], dataset3[0:size]
4.     generated batch[x_fake] = {x : x ∈ generator(a,c) with a,c in x_real1,x_real3}
5.     update the discriminator by descending gradient for real and fake batch sequentially

$$\nabla_{\theta d} = \frac{1}{m} \sigma \left[ \log \log \left( d(x_{n+1}) \right) + \log(1 - d(g(x_n, x_{n+2}))) \right]$$

6.     update the gan using pseudo-huber loss

$$\nabla_{\theta g} \ 0.25 \left( \sqrt{1 + \frac{(d(x_{n+1}) - d(g(x_n, x_{n+2})))}{0.5}} - 1 \right)$$

7.     Calculate ssim score and psim score as

$$SSIM(x,y) = \frac{(2\mu_x\mu_y + c_1)(1\sigma_{xy} + c_2)}{(\mu_x^2 + \mu_y^2 + c_1)(\sigma_x^2 + \sigma_y^2 + c_2)}, PSNR = 10 \log_{10} \frac{\max(y)^2}{\frac{1}{n}\sum_i^n (y_i - \overline{y_i})^2}$$

8.   **end for**
9. **end while**

---

As shown in Fig 1 The discriminator has iteratively trained with true and fake frames alternatively and the score of the discriminator is used to assess its performance, this performance is used to improve adversary performance of the generator. As the discriminator performance improves the generator performance decreases respectively, thus the generator has to further improve itself to defeat the discriminator.

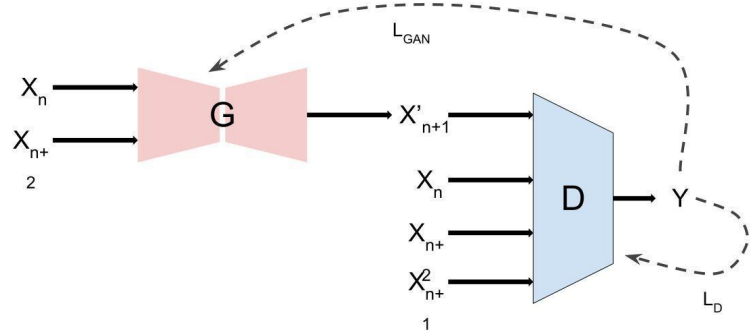

Fig 1. FREGAN model with generator network (G) and discriminator network (D), where G takes as input 2 images $X_n$ and $X_{n+2}$ and discriminator takes as input the same 2 images along with an additional testing image which is either the Xn+1 which is the n+1th frame or X'n+1 which is the predicted n+1th predicted by the generator (G), the discriminator (D) predicts the Fake/True label patch for each pair of image an improves using the discriminator loss of LD and the generator to trains from the prediction of D through Loss LGAN

The model uses 2 loss functions, namely the discriminator loss and the adversarial loss. For the discriminator loss, the proposed approach uses the log loss and for the adversarial loss we have applied the pseudo-Huber loss function for a more robust prediction which can be represented as :

$$L_\delta = \delta^2 \left( \sqrt{1 + \left(\frac{v}{\delta}\right)} - 1 \right) \quad (1)$$

where v is the difference between the actual and the predicted value ie. a = y - $f$(x) making the overall adversarial loss as :

$$L_{GAN} = GD \; \delta^2 \left( \sqrt{1 + \frac{(d(x_{n+1}) - d(g(x_n, x_{n+2})))}{\delta}} - 1 \right) \quad (2)$$

Xn and Xn+2 are the 2 consecutive frames and Xn+1 is the intermediate frame that is to be predicted. D(x) represents the probability of the frame x being real and G(a,b) is the Generators prediction when it predicts an image from noise provided to it along with the 2 frames.

3.1.1 Generator Model

The generator model used in the proposed approach takes 2 subsequent frames and predicts the intermediate frame based on these inputs. The generator consists of an encoding block and decoder block which make it able to predict Xn+1th frame of size 256,256 pixels from an array of 2 frames [Xn, Xn+2], both of size 256,256 pixels.

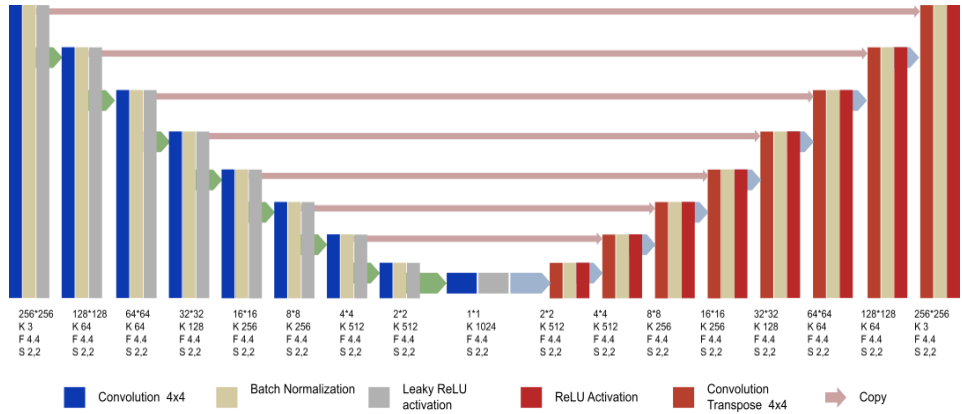

Fig 2. Generator model architecture with 17 layers (9 contraction or encoding layers and 8 expansion or decoding layers)

The encoder block consists of 9 sets of downsampling layers as shown in Fig 2, which include in the order, a 2D convolution operation, a batch normalization operation, and a LeakyReLU activation with convolutional filters increasing gradually. and the compressing nature of these operations gradually decreases the size of the image while increasing the depth.

The decoder block consists of 8 sets of upsampling layers which use 2D convolutional transpose for upsampling operation, which uses the necessary features from the convolution layers in the encoding block by concatenating the output of the convolutional layer with the convolutional transpose for upsampling this block uses the Rectified linear unit as the activation function and outputs a 256,256 3 channel image, which is the predicted frame. The decoder block uncompresses the data passed to them and decreases the depth while increasing the size of the frame.

3.1.2 Discriminator Model

The discriminator model uses a multilayer convolutional neural network to predict whether the input image is fake or real.

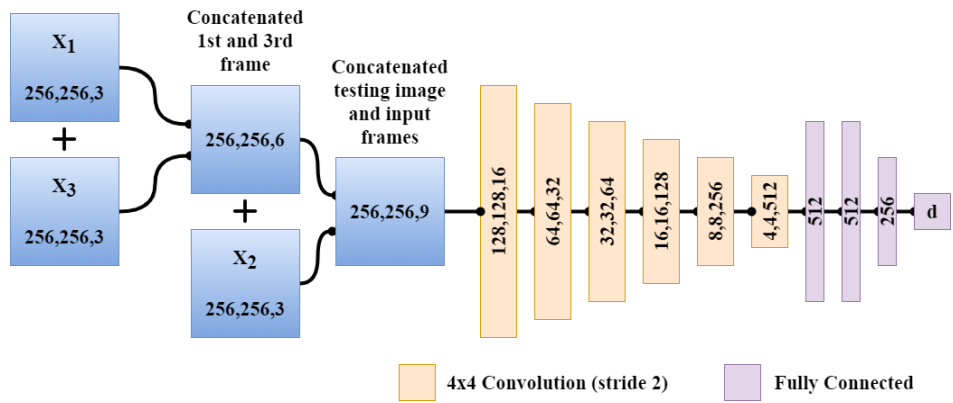

Fig 3. Discriminator model architecture with 10 layers

The generator is accepted as input [$X_n$, $X_{n+2}$] where $X_n$ and $X_{n+2}$ are the nth and n+2nd frames as the source input array and the discriminator takes either $X_{n+1}$, $X'_{n+1}$ as the testing image, where $X_{n+1}$ is the real image and the $X'_{n+1}$ is $G(X_n, X_{n+2})$ the discriminator outputs the log loss, $D(X)$ for real image and $D(X')$ for fake image, due to the log loss the score lies between 0 and 1, with 0 being real and 1 being fake.

# 4 Results and Discussion

In this section, we discussed the dataset, evaluation, implementation, and results. We implemented the FREGAN model on standard datasets that are UCF101[20] and RFree500. For evaluation, we used 2 accuracy metrics that are Structural Similarity Index(SSIM) and Peak Signal to Noise Ratio (PSNR). After this, we implemented the FREGAN model on Keras and TensorFlow. And at last, we discussed the quantitative results of the model.

## 4.1. Dataset

We trained the proposed model with 256 x 256 px 3 channel images which are all frames from 30fps and 24fps videos. We trained on 2 datasets, UCF101[20] and a collection of royalty-free videos scraped from various sources which henceforth we would be calling RFree500, UCF101 is an action recognition dataset with 101 classes and a total of 13320 videos, The RFree500 dataset consists of 500 videos with a varying resolution, all resized to 256 x 256 px, these videos provide a sample of modern recorded videos and were used to test the performance on recreating the details that modern videos contain.

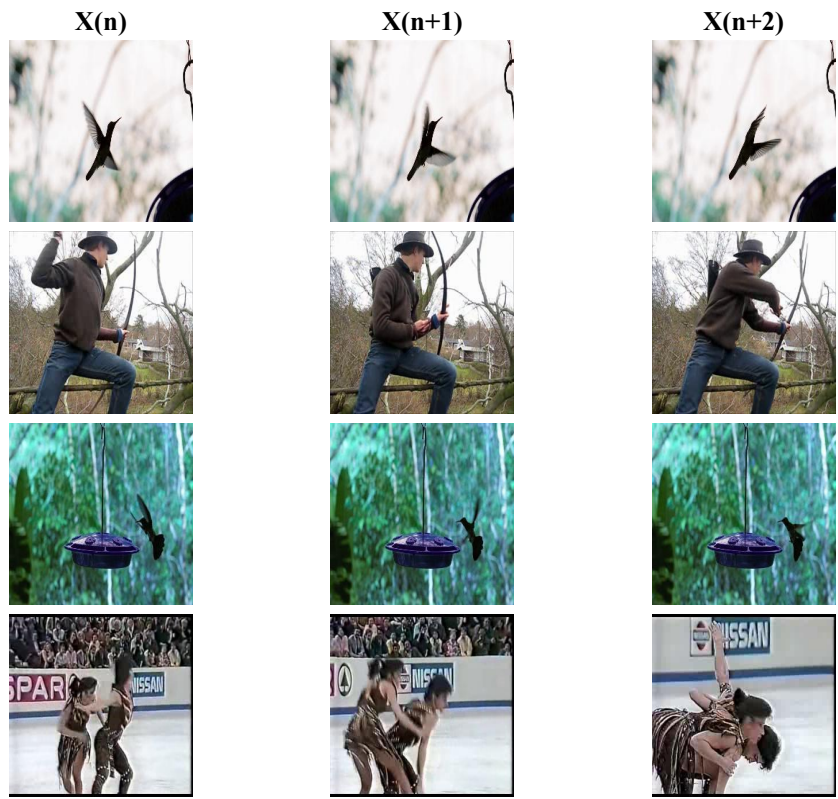

Fig 3. Sample image sequences from the RFree500 dataset and IUC101 dataset which are used for training and testing purposes

### 4.2 Evaluation metrics

For comparison, we have utilized two accuracy metrics: Structural Similarity Index(SSIM) and Peak Signal to Noise Ratio (PSNR). The PSNR and SSIM of the two images can be calculated by equations 3 and 4 respectively :

$$PSNR = 10 \log_{10} \frac{max(y)^2}{\frac{1}{n}\sum_{i}^{n}(y_i - \bar{y}_i)^2} \quad (3)$$

where n refers to the total number of pixels in the image, and y and ỹ are the values of the ith pixel of the real image and generated image

$$SSIM(x, y) = \frac{(2\mu_x \mu_y + c_1)(1\sigma_{xy} + c_2)}{(\mu_x^2 + \mu_y^2 + c_1)(\sigma_x^2 + \sigma_y^2 + c_2)} \quad (4)$$

where μx and μy is the mean of all values of x and y respectively and σx,σy is the variance of those values, σxy being the covariance.

PSNR is actively used in quality assessment for color images as it gives a result that is in line with human perception of the images while SSIM has been traditionally used for the comparison of generated images and original images.

### 4.3 Implementation details

For the proposed experimental setup, we used NVIDIA Tesla T4 GPU with 16GB VRAM along with an intel Xeon 2.0GHz CPU. The model was implemented on Keras with TensorFlow as the backend. We utilized the Adam optimizer[21] for the optimization with beta 1 = 0.00 and beta 2 = 0.95, with the initial learning rate set to 0.0001.

### 4.4 Quantitative results

We experimented with varying the δ threshold in the proposed Huber loss function to evaluate the optimum value of δ for the proposed data and model. This was done by cyclically training the generator and discriminator, the discriminator was trained with both generated images from the iteration of the generator and the target image for 50 epochs for each set of 3 consecutive frames from the dataset. We performed this step for both of the proposed datasets namely, UCF101 and RFree500.

Table 1. Comparison result for varying delta parameters in the proposed loss function for UCF101 and RFree500 Datasets, Showing the highest SSIM and PSNR for both the datasets to be achieved at 0.5

| delta (δ) | RFree500 | | UCF101 | |
|---|---|---|---|---|
| | PSNR | SSIM | PSNR | SSIM |
| 0.1 | 25.26 | 0.82 | 22.58 | 0.69 |
| 0.25 | 24.89 | 0.75 | 20.45 | 0.66 |
| 0.45 | 28.79 | 0.89 | 25.42 | 0.7 |
| 0.5 | **35.19** | **0.97** | **34.22** | **0.95** |
| 0.75 | 26.23 | 0.91 | 27.1 | 0.86 |
| 0.8 | 32.46 | 0.94 | 30.79 | 0.9 |
| 1 | 27.66 | 0.85 | 26.22 | 0.82 |

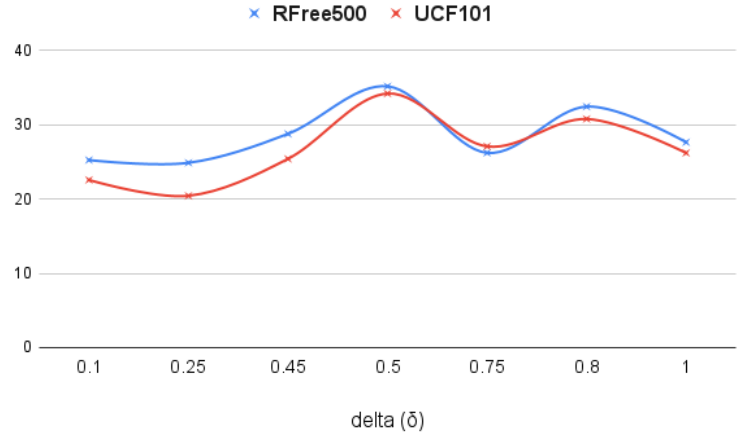

(a) the plot of PSNR and delta values for RFree500 and UCF101 datasets with x-axis representing the delta value and y-axis the PSNR

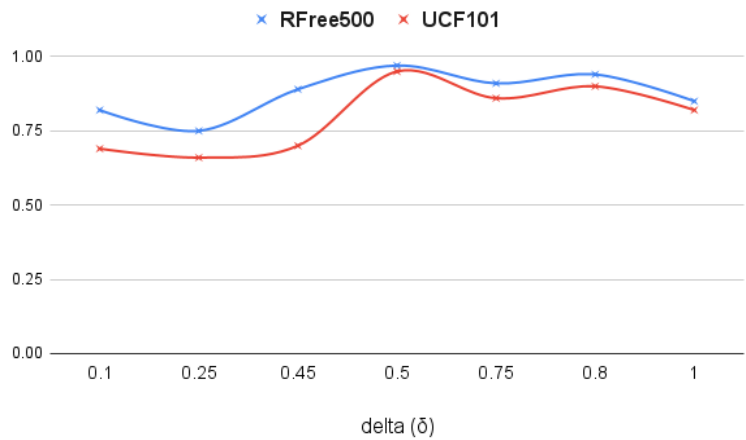

(b) the plot of SSIM and delta values for RFree500 and UCF101 datasets with x-axis representing the delta value and y-axis the SSIM

Fig 4. Visual representation of comparison results for varying the delta parameter in the proposed loss function for UCF101 and RFree500 Datasets, Showing the highest a. (top) PSNR and b.(bottom) SSIM for both the datasets to be achieved at 0.5,(a) and SSIM(b)

From Table 1 it can be seen that the optimum value for the proposed model and data is found to be 0.5, which provides a higher result in both the UCF101 and RFree500 dataset; this data can be visualized better through Fig 4, which shows the same results.

For further experiments we have used the value of δ = 0.5 for the proposed Huber loss function, putting this value of delta in equation (2) we get :

$$L_{GAN} = GD\ 0.25\left(\sqrt{1 + \frac{(d(x_{n+1}) - d(g(x_n, x_{n+2})))}{0.5}} - 1\right) \quad (5)$$

Setting the delta parameter as in equation 5, we again trained the FREGAN model for 15000 steps.

The finalized model after this stage provides a training loss of 0.002 and PSNR of training data as 38.85 while SSIM being 0.98. Testing data which is 13% of the total data gives us a result of 36.49 and 34.94 respectively for the PSNR values of RFree500 and UCF101 datasets.

Table 2. Comparison between previously used methods for frame prediction and frame rate improvement for UCF101, showing that the RCGAN performs better in terms of PSNR while the proposed model has the edge in terms of SSIM

| Methodology | UCF101 | |
| --- | --- | --- |
| | SSIM | PSNR |
| BeyondMSE[22] | 0.92 | 32 |
| ContextVP[23] | 0.92 | 34.9 |
| MCnet+RES[24] | 0.91 | 31 |
| EpicFlow[25] | 0.93 | 31.6 |
| DVF[26] | 0.94 | 33.4 |
| RCGAN[27] | 0.94 | **35** |
| Ours | **0.95** | 34.94 |

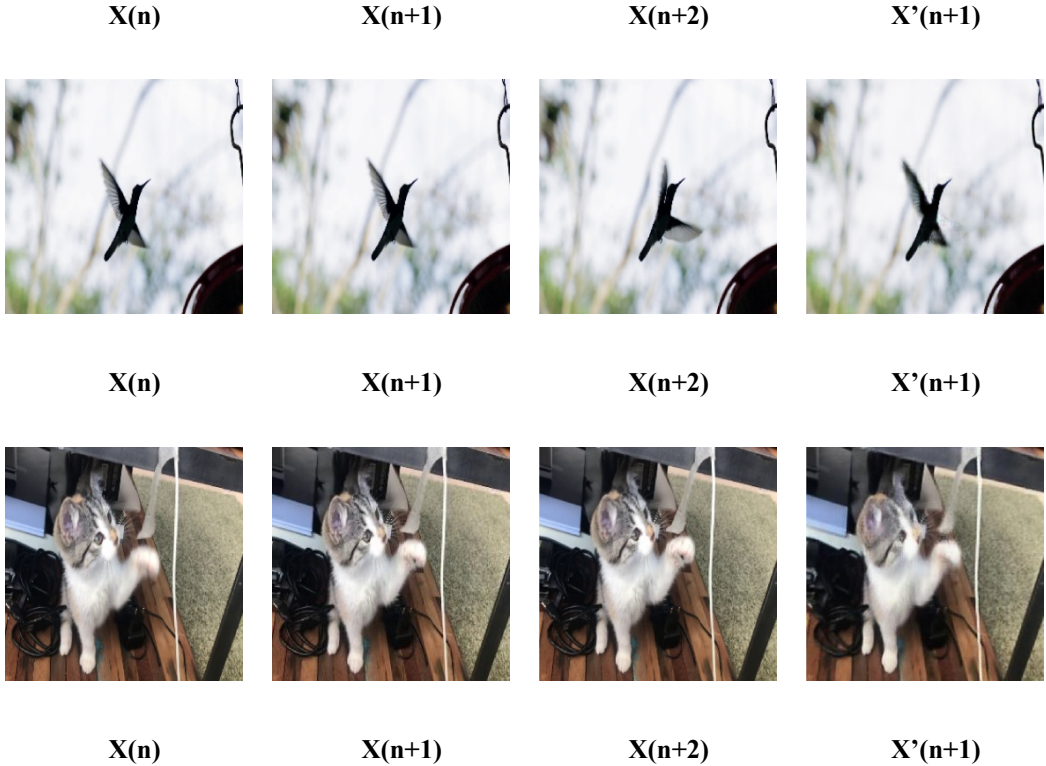

X(n)    X(n+1)    X(n+2)    X'(n+1)

X(n)    X(n+1)    X(n+2)    X'(n+1)

X(n)    X(n+1)    X(n+2)    X'(n+1)

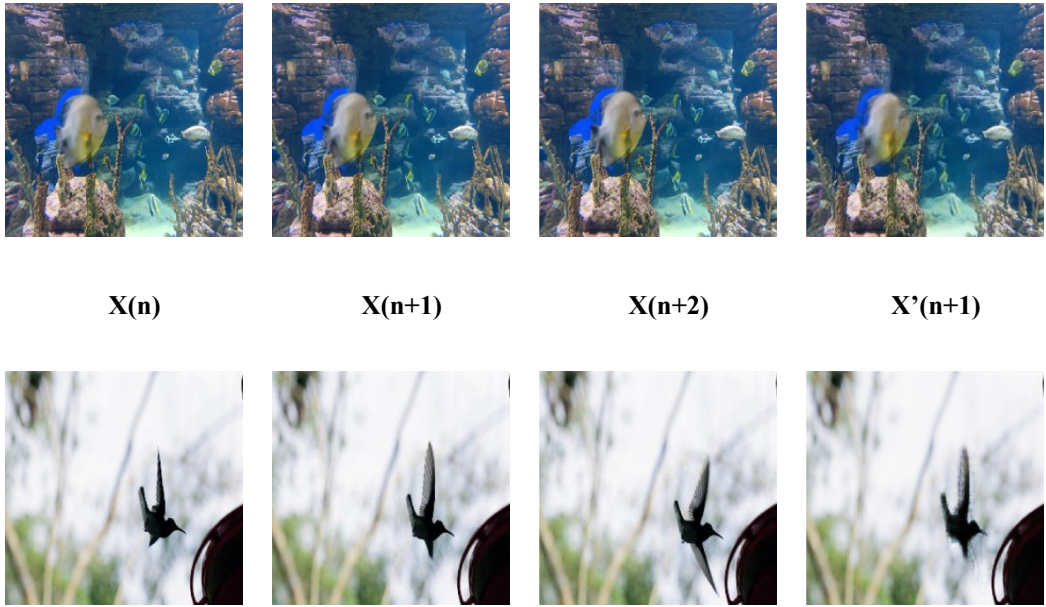

| X(n) | X(n+1) | X(n+2) | X'(n+1) |

Fig 5. The output of the proposed trained FREGAN model showing results from various frames in the RFree500 dataset

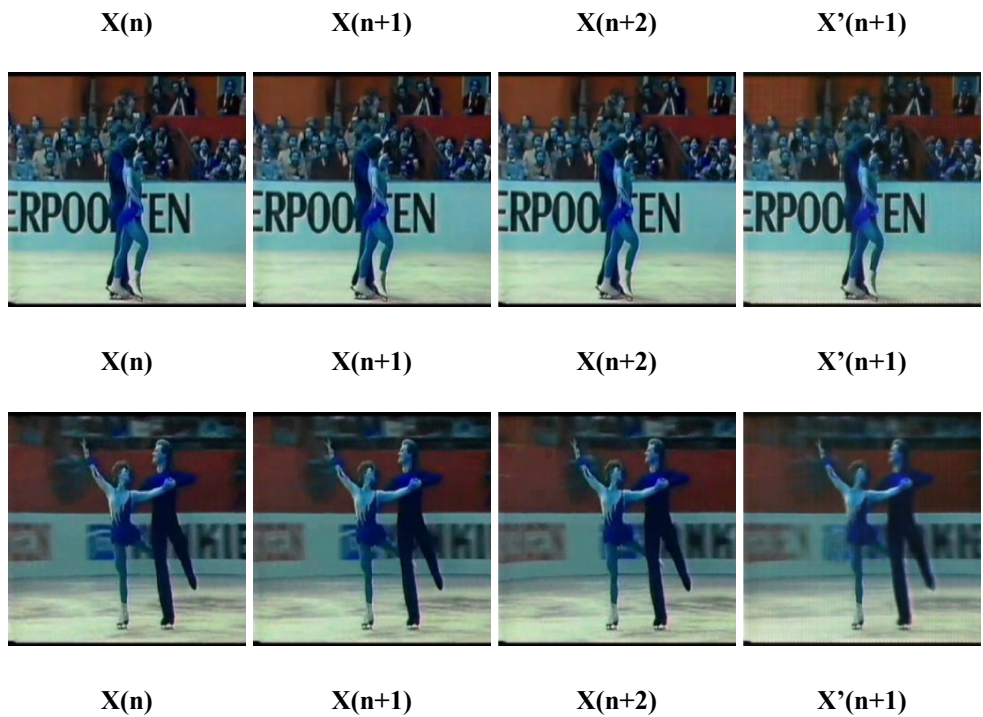

| X(n) | X(n+1) | X(n+2) | X'(n+1) |

| X(n) | X(n+1) | X(n+2) | X'(n+1) |

| X(n) | X(n+1) | X(n+2) | X'(n+1) |

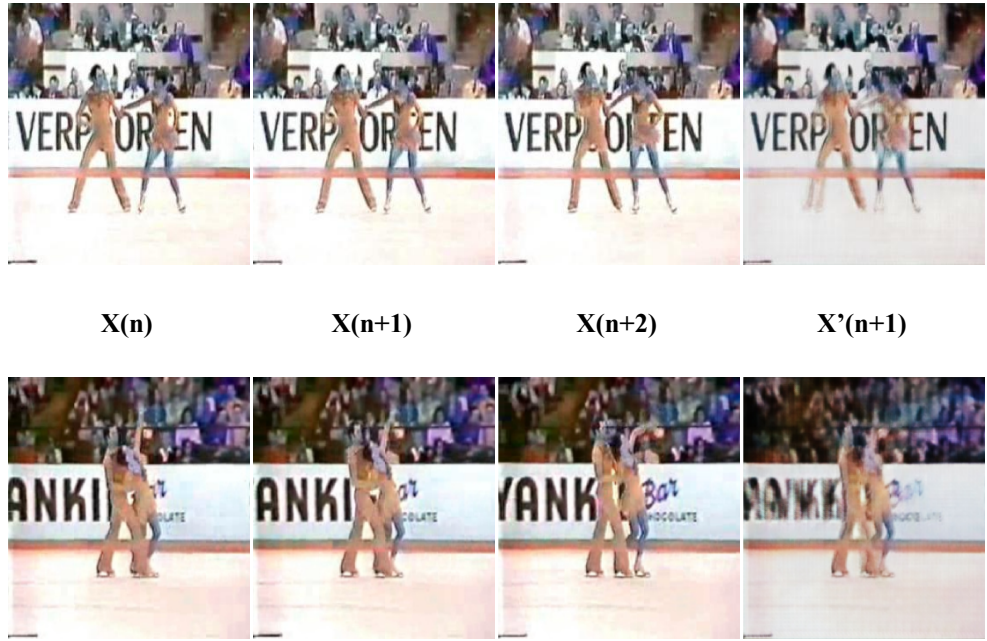

| X(n) | X(n+1) | X(n+2) | X'(n+1) |

Fig 6. The output of the proposed trained FREGAN model showing results from various frames in the IUC101 dataset

### 4.5 Discussion

Comparing the proposed model with other published methods for the task of image frame rate enhancement as shown in table 2, shows that the proposed method resulted in a higher SSIM score than the others and a PSNR score only 0.0017% lower than the best performing method for the UCF101 dataset. The consistency across the images shown in Fig 5 and Fig 6 can be attributed to using a very similar discriminator to the PatchGAN, which only penalizes the local image patches convolutionally instead of the entire image, this helps in cases of images where the foreground or smaller clustered regions show changes while the general background remains constant, in such cases the discriminator penalty would be distributed accordingly while the Huber loss in the GAN allows it to effectively predict the results in the case of UCF101, as owing to Huber loss being quadratic under delta and liner over the delta, this allows FREGAN to successfully work with noise in an image effectively reducing the effect of outliers or, noise in the proposed case, from affecting the scores of the proposed model excessively.

### 5. Conclusion

Comparing the proposed model with other state-of-the-art methods for the task of image frame rate enhancement as shown in table 2, we can conclude the high effectivity of Huber loss in the performance of the proposed GAN model. The model size being smaller than many methods gives the proposed model a chance to be used in high-efficiency video frame rate enhancement. Future work in improving the generator structure further and increasing the number of intermediate frames predicted can be undertaken with only minor modifications to the model and choosing a higher quality image dataset could improve the performance in the above said task.